# Self-Evolutionary Reservoir Computer Based on Kuramoto Model

Zhihao Zuo[1,3], Zhongxue Gan[1,2,a)], Yuchuan Fan[4], Vjaceslavs Bobrovs[5], Xiaodan Pang[3,4,5], Oskars Ozolins[3,4,5,a)]

[1]Academy for Engineering and Technology, Fudan University, Shanghai 200433, China

[2]Department of Engineering Research Center for Intelligent Robotics, Jihua Laboratory, Guangdong 528200, China

[3]Department of Applied Physics, KTH Royal Institute of Technology, 106 91 Stockholm, Sweden

[4]RISE Research Institutes of Sweden, 164 40 Kista, Sweden

[5]Institute of Telecommunications, Riga Technical University, Riga, Latvia, LV-1048

a) Author to whom correspondence should be addressed: ganzhongxue@fudan.edu.cn, ozolins@kth.se

**Abstract**: The human brain's synapses have remarkable activity-dependent plasticity, where the connectivity patterns of neurons change dramatically, relying on neuronal activities. As a biologically inspired neural network, reservoir computing (RC) has unique advantages in processing spatiotemporal information. However, typical reservoir architectures only take static random networks into account or consider the dynamics of neurons and connectivity separately. In this paper, we propose a structural autonomous development reservoir computing model (sad-RC), which structure can adapt to the specific problem at hand without any human expert knowledge. Specifically, we implement the reservoir by adaptive networks of phase oscillators, a commonly used model for synaptic plasticity in biological neural networks. In this co-evolving dynamic system, the dynamics of nodes and coupling weights in the reservoir constantly interact and evolve together when disturbed by external inputs.

## I. Introduction

Neuromorphic computing, involving brain-inspired computing architectures, can mimic the behaviors of the nerve systems to solve challenging problems in the machine learning (ML) community.[1] As a neuromorphic computing paradigm, reservoir computing (RC) has incomparable advantages in extracting spatiotemporal information from real-world problems,[2–4] such as speech recognition,[5,6] dynamic pattern recognition,[7,8] financial forecasting,[9] and robot motor control.[10] Typically, RC consists of many interacting neuronal nodes, termed the "reservoir." Even such a simple network can generate complex dynamics.[11] When an external input signal is fed, the dynamics of the reservoir can draw out spatiotemporal information and map it to a high-dimensional feature space. This dynamic system contains plenty of input data features, like a reservoir. Then only the simple readout weights are trained to read the state of the reservoir and mapped to the desired output.[2,12] That is, once the perturbation induces the transient dynamics in the reservoir, the readout layer translates the traces of the system to the target output. Compared with other recurrent neural networks (RNNs), RC training is only performed on the readout layer, leading to a more efficient training phase.[13,14]

Conventionally, the typical reservoir employs fixed random connections, which is considered the main advantage of RC.[2] Intuitively, the random structure of the reservoir is unlikely to be optimal. Accumulating evidence indicates that the reservoirs only show the best performance "at the edge of chaos."[11,15] However, the probability of exactly hitting the minimum in the error landscape of RC by picking a random point is virtually equal to zero. So, the specification of the reservoir requires numerous trials and even luck.[16] Several theories on RC have been reported. For example, Jaeger proposed the spectral radius to be slightly lower than one to guarantee the reservoir's echo state property (ESP).[17] Furthermore, a tighter bound on ESP is also described.[18] However, how to construct reservoirs with certain 'desirable' properties for a specific application in practice remains to be determined.

Some efforts have been made to alter the reservoirs to improve performance on a given application. First, we can use some search algorithms in the ML community to find an optimal reservoir. Such search algorithms, e.g., Genetic Algorithms,[19] NEAT,[20] and Evolino,[21] are based on the fact that the performance of random reservoirs forms a distribution. Second, the idea of pruning away "bad" nodes from a large initial reservoir was also proposed.[22] Lastly, another prevailing proposal is reservoir adaptation. The reservoir parameters are changed using the unsupervised learning rule in such a scheme. The external dynamics can be embedded in the reservoir without explicitly expected response.[20,23] Moreover, the sub-reservoirs can also be taken as the smallest adaptation units rather than neuron nodes.[24,25] Eventually, multiple sub-reservoir networks are constructed. These algorithms can easily perform by choosing the appropriate structure. However, inherent defects in ML are introduced, such as black box, high computation cost, and slow convergence, which counter RC's nature. Besides, some biologically inspired generic methods are also presented for creating RC with different neuron models, such as the Hebbian rule, Intrinsic Plasticity.[26–29] However, per Jaeger, simply imposing Hebbian rules on RC cannot significantly improve performance.[30] The simple correlation or anti-correlation from Hebbian rules is too limited to improve the reservoir dynamics.

Typical reservoir constructions only consider static networks or separate the dynamics of neurons and connectivity. There are approximately 86 billion neurons in the typical human brain. When we look at the brain, a specific structure is present, which is essential for information processing.[31] Microscopically, synapses in the brain have great activity-dependent plasticity. The connectivity patterns of neurons change dramatically depending on the activities of neurons.[32] Macroscopically, brain structural changes reveal nonlinear trajectories with the development of neurons over the lifespan.[33] It has been reported that the evolution of the network topology is significantly affected by the states of the elements in many biological networks and vice versa.[32] Adaptive networks are commonly used for synaptic plasticity, which determines learning, memory, and development in neural networks.[34,35] In this co-evolving dynamic system, the coupling weights between nodes and the states of the active elements at the nodes interact and evolve together. Explosive synchronization (ES) is more popular in adaptive networks, where the reservoir maintains excellent computing performance.[36,37]

Inspired by these biological and physical findings, we propose a structural autonomous development reservoir computing model (sad-RC) to implement structural and connectivity plasticity in the reservoir. In the development stage, with the stimulations of external inputs, the dynamics of nodes and synapses in the reservoir constantly evolve together. Such an autonomous development mechanism makes the reservoir adapt its internal dynamics to the given task. Consequently, a specific structure reservoir is developed with the "physics plus information."[38] To the best of the authors' knowledge, most current studies focus on the evolution of networks from generation to generation. At the same time, the development of the reservoir structure with external stimulations has not yet received sufficient attention.

In this paper, we first evaluate the performance of the sad-RC in four representative artificial and real-life dynamics on two benchmark tests: prediction and memory capacity. Then, the order parameter is introduced to interpret the generalization capability of the model over a vast domain of error landscape. Subsequently, we try to answer the puzzle[30] why simply applying Hebbian rules cannot significantly improve the reservoir performance from the synchronization perspective. Both experiments and theoretical analyses indicate that our model can simulate many development mechanisms of neuron networks in vivo. Our model gives a simple yet powerful interpretation of how neural networks in the human brain develop certain connectivity patterns without well-understood unsupervised adaptation. Therefore, it is a good candidate for next-generation RC and designing neuromorphic hardware.

## II. Methodology

### 2.1 Structural Autonomous Development Reservoir Computer

We consider an adaptive network of $N$ oscillators perturbed by external inputs, in which the coupling weights between the nodes and the state of the active elements at the nodes interact and evolve together. When data is fed into the reservoir computer, the reservoir nodes draw information and transform it from low-dimensional input space to high-dimensional feature space. Simultaneously, the connections adjust their coupling strength automatically according to the autonomous development mechanism. The basic idea underlying the structure of the self-adaptive oscillatory reservoir computer is: if all the necessary information to construct proper computational results was provided, not only the trajectories of the reservoir state should converge under the governing of ESP, but also the structure of the reservoir internal connection should evolve to a task-specific network controlled by autonomous development mechanism.

If all the necessary information to construct proper computational results was provided, the trajectories of the reservoir state should converge under the governing of ESP. Meanwhile, the reservoir's internal connection structure can evolve into a task-specific network controlled by an autonomous development mechanism.

We begin with considering a network of N Kuramoto-like phase oscillators. The dynamics of each of them can be described by a phase $\theta_i(t) \in [0, 2\pi)$

$$\frac{d\theta_i}{dt} = \omega_i + \lambda \sum_{j=1}^{N} k_{ij} \sin(\theta_j - \theta_i + u(t)), \quad (1)$$

where $\omega_i$ is the natural frequency of the node $i$, $i = 1, \dots, N$, $\lambda > 0$ is the global coupling strength, and $k_{ij}$ is the entry of the adjacency matrix $K$ of the network. $u(t)$ is the input data. We use this extended Kuramoto model to simulate neuronal behavior in the reservoir computer pool.[39]

Next, we propose a dynamical adaptive model for the coupling weights $k_{ij}$. Because it is natural that their dynamics only depend on the relative timing of the oscillators, the following is such a reasonable model

$$\frac{dk_{ij}}{dt} = -\varepsilon \sin(\theta_j - \theta_i + \beta), |k_{ij}| \leq 1. \quad (2)$$

The parameter $\varepsilon$ represents the time scale of this dynamics, which is much longer than that of the dynamics of the oscillators, i.e., $\varepsilon \ll 1$. The constraint condition $|k_{ij}| \leq 1$ implies that if $k_{ij}$ takes a value outside the interval $[-1,1]$, it is immediately set to the appropriate limiting value (-1 or 1).

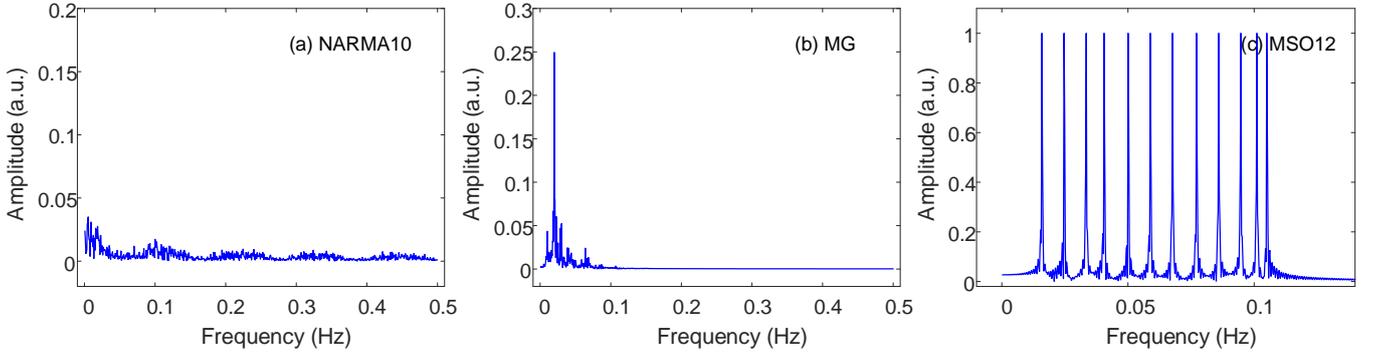

Fig. 1. Fast Fourier Transform results of the input signal of (a) NARMA10, (b) MG time series, and (c) MSO12.

Because the coupling weight cannot grow indefinitely, this limitation is reasonable. We numerically confirmed that even though the distribution of coupling weights is randomly initialized, the network will soon converge to a specific structure with the perturbation of the input stream. This development function determines how the changes in the coupling weights depend on the phase differences among the oscillators. Here, we term $\beta$ character parameter because it controls the characteristics of the development function, which we also make a sophisticated study in the latter.

According to the traditional training method, the reservoir is initialized randomly. Before reaching a stable neuronal state, the first 100 to 200 steps of neuronal states are discarded conventionally, known as the washout phase.[17] In many scenarios, data is too expensive to get more. Discarding data arouses the data waste problem. To fully use the data, we emphasize that the neural network can evolve and adjust synaptic weights distribution with the external stimulation in the washout phase, eventually reshaping a stable structure. Algorithm 1 displays the autonomous development mechanism procedure used by the reservoir computer:

Where $\hat{y}(t) \in \mathbb{R}^{N_Y}$ denotes the output at the time $t$, and $W_{out} \in \mathbb{R}^{N_Y \times N_R}$ is the reservoir-to-output weight matrix. For simplicity, we omit the input layer and input weights $W_{in}$ in this model. Contrary to the fixed adjacency matrix $W_{res}$ in the traditional reservoir, the neuron synapses can adjust their synaptic weights distribution autonomously by the autonomous development mechanism. Firstly, we evaluate the performance of the structure self-adaptive reservoir computer in some benchmark tasks. Subsequently, we explore why the autonomous development mechanism can bring such dominant advantages through numerical experiments. At last, we compare synaptic weights distribution in the evolved reservoir with real biological neural networks.

### 2.2 Benchmark Tests and Experimental Settings

The reservoir in the RC system can nonlinearly map the sequential input signals onto the high dimensional state space, then process by the readout network using a linear combination of the reservoir node states. Therefore, we can evaluate the quality of the reservoir in two aspects: 1) the quality of nonlinear transformation the reservoir can map; 2) the diversity of the temporal dynamics that the reservoir can capture. We evaluate structure self-adaptive RC systems with different initial architectures on three different time series data: 10th-order nonlinear autoregressive moving average (NARMA10), Mackey-Glass (MG) time series, and multiple superimposed oscillators (MSO). Benchmark task memory capacity is also conducted. Besides, we applied our model to real-life dynamics of sea clutter data. Due to space limitations, the results were included in the supplementary materials.

**NARMA System**

The NARMA is a discrete-time system. The current output depends on both the input and the previous output. The NRMA10 system is quite challenging to model due to its

**Algorithm 1.** The reservoir network evolves to a specific structure under the governance of autonomous development mechanism

1. Initialize the adjacency matrix $K$, neuronal states $\Theta$, spectral radius $\rho$;
2. **For** $i = 1$ to $L_{train}$ (the length of the training data) **do**
3. Update the neuronal states $\Theta$ according to (1);
4. **if** $i < L_{adev}$ (the data length for adapt network structure) **then**
5.     Update adjacency matrix $K$ according to (2);
6.     Rescale spectral radius of matrix $K$: $\rho(K) = \rho$;
7. **Else**
8.     Collect the neuronal states $\Theta$;
9. **End if**
10. **End for**
11. Compete the output weight $W_{out}$ by ridge regression;
12. **For** $i = 1$ to $L_{test}$ (the data length for test model) **do**
13. Update the neuronal states $\Theta$ according to (1);
14. Collect prediction value according to $W_{out}$: $\hat{y}(t) = W_{out}\Theta(t)$;
15. **End for**
16. Calculate the model error.

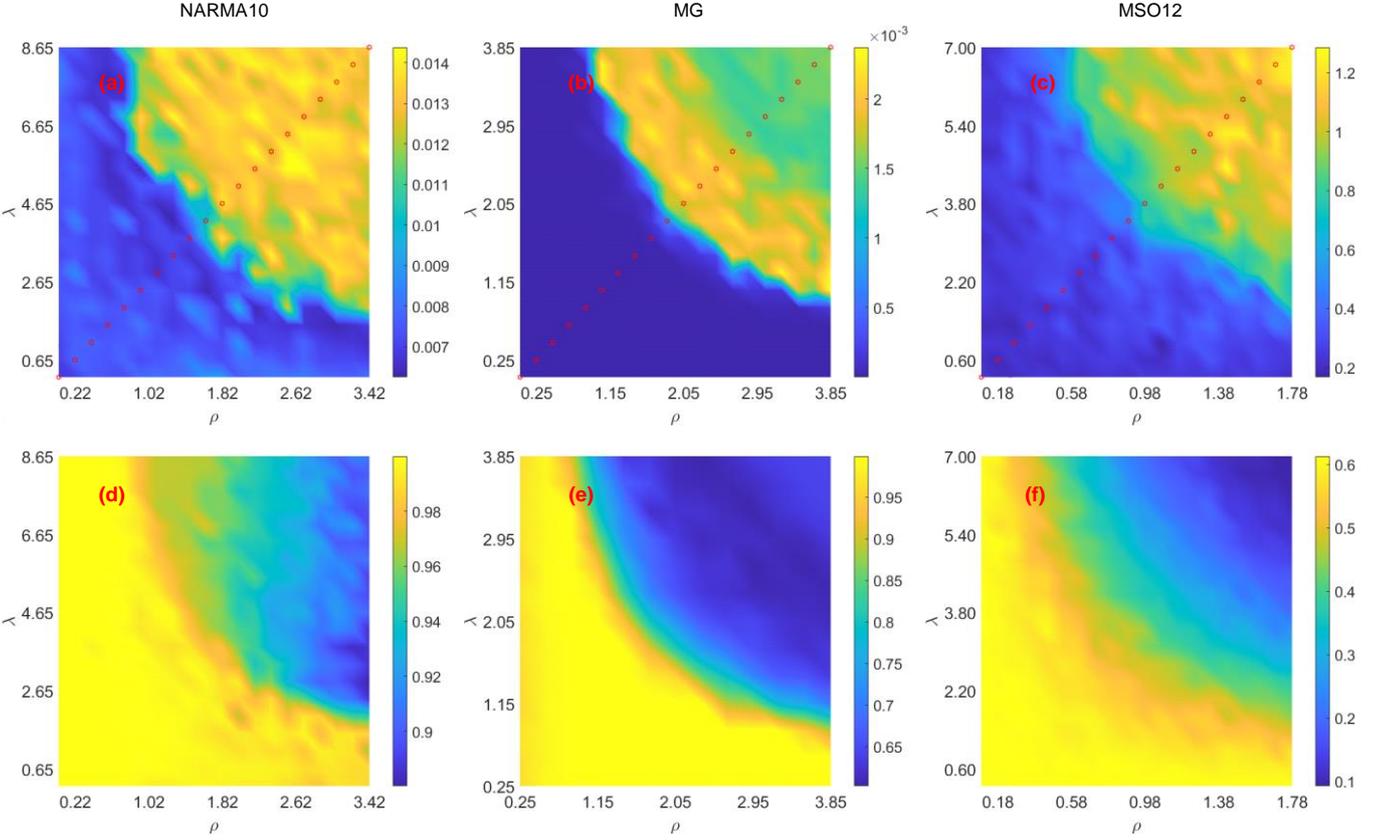

Fig. 2. Test error according to the global coupling strength $\lambda$ and spectral radius $\rho$ in (a) NARMA10, (b) MG, and (c) MSO12. (e)–(f): the measure of synchrony in reservoir nodes corresponding to task a, b and c respectively.

nonlinearity and long memory, which is usually used to test the performance of RNNs.[16,40] The NARMA10 system is generated as follows:

$$y(t+1) = 0.3y(t) + 0.05y(t)\left[\sum_{i=0}^{9} y(t-i)\right] + 1.5u(t-9)u(t) + 0.1. \quad (3)$$

Where $u(t)$ is a random input at the time step $t$, generated from a uniform distribution over $[0,0.5]$ and $y(t)$ is the output at time step $t$. The readout network is trained to predict the signal $y(t+1)$ from the reservoir state.

**MG System**

As a time-series benchmark with a chaotic attractor, the MG has been widely used in the literature to test the performance of a model on dynamical system identification.[41] The MG is derived from a time-delay differential system with the form

$$\frac{dy(t)}{dt} = \frac{ay(t-\tau)}{1+y^n(t-\tau)} + by(t). \quad (4)$$

Here, $y(t)$ is the output at time step $t$, and $\tau$ is the time delay. According to 17 and 41, the parameters of the equation are defined as $n = 10$, and $b = -0.1$. When $\tau > 16.8$, the system has a chaotic attractor. Thus, in this paper, we set $\tau = 17$, and define the task as to perform one-step-ahead direct prediction, $i.e.$, the readout network is trained to predict $y(t+1)$ from the reservoir state when $y(t)$ is given. The equation is solved by the fourth-order Runge-Kutta method with random initialization.

**MSO12 Task**

In contrast to the MG and NARMA system, the MSO time-series prediction is challenging for RC, even though the function consists of only two sine waves. Another important property of the structure self-adaptive RC is the excellent ability to capture diverse temporal dynamics with different time scales. The MSO task is used to analyze the multiple time scales processing ability of the different reservoir structures.[25] The MSO time series is a sum of the sinusoidal functions:

$$u(t) = \sum_{i=1}^{m} sin(\varphi_i t), \quad (5)$$

where $m$ is the number of sine waves, $\varphi_i$ specifies the frequency of the $i$th sine wave. In this paper, the MSO with $m = 12$ (MSO12) is used to test the model's performance. The $\varphi_i$ coefficients are set as in Ref. 42.

As shown in Figure 1(a), (b), and (c), the frequency spectra of the input signal of NARMA10, MG, and MSO12 are calculated by the Fast Fourier Transformation (FFT), respectively. Even though the output of NARMA10 is governed by the input signal base on equation (3), the frequency spectrum of the input signal of NARMA10 has weak frequency dependence. In contrast, MG has clear frequency peaks since the input is the previous output in this task. Notably, the major peaks of the MG time series are located near the low-frequency range rather than spreading out evenly, which is the case of the MSO12 time series.

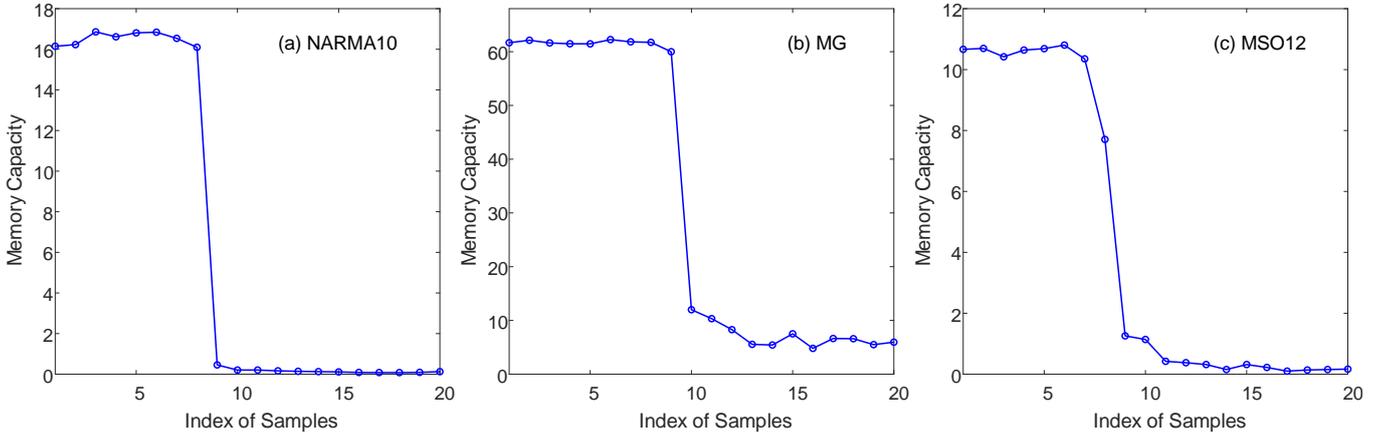

Fig. 3. The memory capacity of the sad-RC in task (a)–(c). The sampling nodes comes from Fig. 2(a)–(c) accordingly.

**Sea Clutter Radar Returns**

We also test the performance of the model on a noisy real-life prediction task. Sea clutter data, collected by the McMaster University IPIX radar, is the radar backscatter from an ocean surface.[43] In this experiment, $y(t)$ is used to predict the next value $y(t + 1)$. The results show in the appendix.

**Experimental Settings**

In the following numerical examples, we test the learning ability of the structure self-adaptive RC. We first keep a single-layer reservoir and fix the number of neuronal nodes in the reservoir $m = 100$. The performance of an RC system is strongly affected by the size of the reservoir network. For each dataset, we denote the length of the autonomous development, training, and test sequence by $L_{adev}$, $L_{train}$, and $L_{test}$, respectively. Table I illustrates the hyperparameters for each task. We emphasize that there are no values from autonomous development. Training and test sequences are used during the initial washout period.

## III. Numerical Test

In this section, we will first compare the performance of the two different reservoir structures on time series prediction tasks and benchmark task of memory capacity (MC), $i.e.$, reservoir computer with (RC) and without (sad-RC) autonomous development mechanism. Then, we analyze the effects of the autonomous development mechanism on the properties of the RC system.

### 3.1 Time Series Prediction Tasks

We first evaluate the performance of the sad-RC on time series prediction tasks in these data sets. We set $\varepsilon = 0.1$, and initialize $\Theta = [\theta_1, \theta_2, \ldots, \theta_{100}] = \vec{0}$. $\Omega = [\omega_1, \omega_2, \ldots, \omega_{100}]$ comes from a standard normal distribution. The density $s$ of the adjacency matrix $K$ is 0.05, $i.e.$, only 5% of entries in $K$ are nonzero, and the others are all zero. In this study, we consider a weighted connection, which means $k_{ij} \in [-1,1]$ rather than a 0 and 1 network. The nonzero entries $k_{ij}$ in $K$ are initialized uniformly from an interval $[-1,1]$. We solve Equations (1) and (2) by the Euler method in each iteration and set timestep $\Delta t = 1.0$. Here, we set character parameter $\beta = \pi/2$ for brevity, and sophisticatedly study the effect of character parameter $\beta$ in the following.

TABLE I. The hyperparameters for tasks.

| Task | $L_{adev}$ | $L_{train}$ | $L_{test}$ | $\lambda$ |
|---|---|---|---|---|
| NARMA10 | 100 | 900 | 500 | 4.0 |
| MG | 100 | 2900 | 1000 | 1.0 |
| MSO12 | 100 | 1200 | 100 | 4.0 |
| Sea clutter | 100 | 1700 | 500 | 0.5 |

When RCs are applied to time-series prediction tasks, in classical RC literature, the first $L_{ini}$ values are used as the initial period to wash out the initial state of the reservoir. In contrast, when the data stream was fed into the sad-RC, the first $L_{adev}$ values are used to shape the structure of the reservoir rather than washout. At the micro level, the synapses adjust synaptic weights distribution under the government of equation (2). At the mesoscopic level, the neural networks in the reservoir continually evolve with the external stimuli. Eventually, ruled by an autonomous development mechanism, a task-tailored neural network in the reservoir is reshaped. The specific development mechanism is controlled by character parameter $\beta$.

After the specific reservoir network evolved from a random initial adjacency matrix, the $L_{train}$ values are fed into the reservoir to train the output weights for the prediction task. The mean-square error (MSE) is used to train and evaluate the performance of the sad-RC

$$MSE = \frac{1}{L_{train}} \sum_{i=1}^{L_{train}} \|y(t_i) - \hat{y}(t_i)\|^2, \qquad (6)$$

where $\hat{y}(t_i)$ is the readout output, $y(t_i)$ is the desired output (target), $\|\bullet\|$ denotes the Euclidean distance (or norm). Here, the effect of two key parameters on the performance of the sad-RC in prediction tasks are carefully studied, $i.e.$, the global coupling strength $\lambda$, and spectral radius $\rho$. The $\lambda$ can have a significant influence on the adaptive oscillator networks, and the $\rho$ plays an essential role in the RC system.[2] For each dataset, keeping the hyperparameters intact, we generate 10 randomly initialized sad-RCs and calculate their average prediction performance.

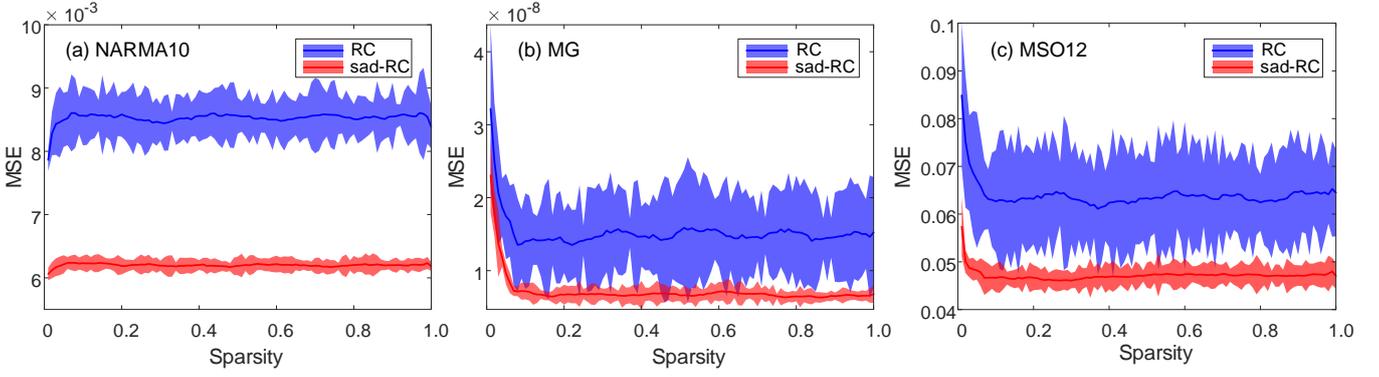

Fig. 4. MSE of test errors vs sparsity in the reservoir. The mean and variance of test errors for RC and sad-RC from 50 simulations in each connection sparsity.

Figures 2(a)-(c) report the averaged errors in the benchmark NARMA10, MG, and MSO12. We measure the errors brought by the change of global coupling strength $\lambda$ and spectral radius $\rho$. It is observed that, in all tasks, the errors are minimized at a vast domain, rather than a critical point in a line or a critical line in a landscape, which is usually reported in general and extended RC model.[11] Besides, the success in MSO12 indicates that the sad-RC can capture multiple time scale in diverse temporal dynamics very well.

### 3.2 Memory Capacity

Then we evaluate the short-term-memory (STM) performance of the sad-RC model on the MC benchmark tasks introduced by Jaeger.[44] The STM is a measure of how much information contained in the input can be extracted from the reservoir after a certain time. In particular, assume that the reservoir is driven by a temporal signal $s(t)$. For a given delay $k$, we consider the reservoir with optimal parameters for the task of reproducing $s(t-k)$ after seeing the input stream $s(t-1)s(t)$ up to time $t$. The performance is measured in terms of the squared correlation coefficient between the observed reservoir output $y(t)$ and the desired output (input signal delayed by $k$ time steps)

$$MC_k = \frac{Cov^2(s(t-k),y(t))}{Var(s(t))Var(y(t))}, \quad (7)$$

where $Cov$ denotes the covariance and $Var$ the variance operators. The STM capacity is then defined as

$$MC = \sum_{k=1}^{\infty} MC_k. \quad (8)$$

We set the maximum time delay $k$ up to 100 in practical implementation. The input signal $s(t)$ is drawn from a uniform distribution in the interval $[-0.5, 0.5]$. The sampling nodes are marked by a red pentagon in Fig. 2(a)-(c). In this study, the MC of the sad-RC is evaluated after it completes the prediction task in these sampling nodes with intact hyperparameters. As Fig. 3 shows, the sampling nodes orderly come from low right to upper right in Fig. 2(a)-(c) accordingly. With the increase of global coupling strength $\lambda$ and spectral radius $\rho$, the memory capacity keeps in a high plateau, then undergoes a sharp decline. In three tasks, both the high and low MC in Fig. 3 coincide with the great and worse performance in Fig. 2(a)-(c), respectively, which indicates the high MC and great prediction performance for the reservoir are not separatable. We conclude that our model can show great performance in a vast domain in both prediction tasks and MC tests. According to the embedding and approximation theorems,[45] the trained RC can exhibit dynamics that are topologically conjugate to the future behavior of the observed dynamical system. Consequently, excellent memory capacity provides the basis for high prediction accuracy in the prediction stage.

### 3.3 The Effect of Autonomous Development Mechanism

To find out why the sad-RC shows excellent performance in the vast domain, we introduce a measure of synchrony to explore the effect of the autonomous development mechanism. One measure of synchrony is the Kuramoto order parameter

$$re^{i\theta} = \frac{1}{N}\sum_{j=1}^{N} e^{i\theta_j}. \quad (9)$$

The order parameter $r$ achieves its maximum 1 when the phase of all oscillators is identical in complete phase synchronization. It becomes close to 0 when the phases are scattered around the circle in dynamical incoherence.

We measure the synchrony of reservoir nodes after they complete the autonomous development stage. As shown in Fig. 2(d)-(f), the landscapes are divided into yellow and blue areas separately, which can perfectly match their counterparts in Fig. 2(a)-(c), respectively. The order parameter $r$ keeps in a high plateau but is not equal to 1, which is like the state of "the edge of chaos" reported in Ref. 11. Combining with the discovery that the RC shows the best performance at the edge of chaos,[15] we conclude that the autonomous development mechanism can broaden the state of "the edge of chaos" from a critical point or line to a vast critical domain, where the sad-RC shows excellent performance in prediction tasks and MC tests.

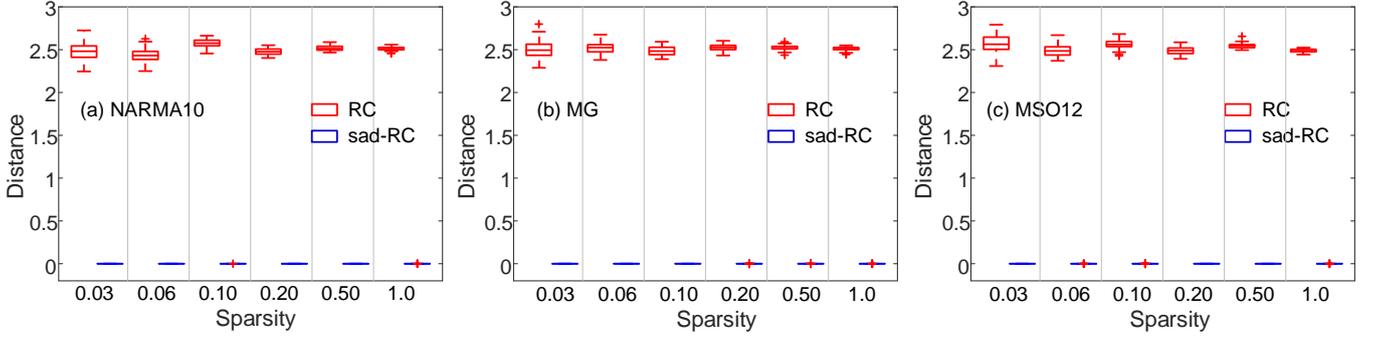

Fig. 5. The distance of adjacency matrices from 50 independent simulations in RC and sad-RC for each sparsity in (a) NARMA0, (b) MG and (c) MSO12 respectively.

The autonomous development mechanism can persist reservoir in a critical state at a broad parameter domain. In adaptive oscillator networks, the adaption mechanism draws oscillators into complete synchronization when parameters are within the collapse threshold. Thus, we attribute that the autonomous development mechanism promotes the neurons in the reservoir to complete synchronization. However, with the constant disturbances of the input stream, only the imperfect synchronization state can be persisted, $i.e.$, the state of "the edge of chaos." When parameters exceed the threshold, the adaption mechanism fails, and the neurons burst into a mass of tangle, which is supported by order parameters in Fig. 2.

### 3.4 Sensitivity and Effectiveness Analysis

We test the sensitivity of the model performance on one-step ahead prediction concerning variations in the construction parameter. Since the performance of RC is detrimentally affected by the structure of the reservoir, in this paper, we keep other hyperparameters identical and vary the sparsity of the initial adjacency matrix from no connection ($s = 0$) to full connection ($s = 1$) in the RC and sad-RC respectively. For each density, 50 independent adjacency matrices are generated for reservoirs. The mean and variance of test errors from RC and sad-RC are plotted in Fig. 4. It is observed that both types of reservoirs display better and more stable performance with increasing connection density. Besides, the mean and variance in the sad-RC are lower than RC. Thus, we can conclude that the autonomous development mechanism is conducive to RC showing better and more robust performance.

### 3.5 Astringency

The performance of RC is largely determined by the adjacency matrix. To answer why the sad-RC enjoys robustness, we study the astringency of the autonomous development mechanism. In a given connection sparsity, 50 initial connection matrices are generated. The distance of connection matrices between $K_i$ and $K_1$ is evaluated by $d_i = \sum_{j=1}^{100}\sum_{k}^{100}(K_{jk}^i - K_{jk}^1)$, After employed to sad-RC, 50 specific adjacency matrices are achieved $K_1', K_2', \ldots K_{50}'$. We calculate the distance between the matrices, $d_i'$, as superior.

Without loss of generality, we vary connection sparsity from 0.03 to 1.0 and set the character parameter $\beta = 0$. As shown in the boxplot in Fig. 5, the median value of the distance of connection matrices in the traditional reservoir from 50 independent simulations is about 2.5. However, the counterpart in sad-RC is approximate $1.0 * e^{-16}$. The sad-RC enjoys a great astringency which is the fundamental of robustness.

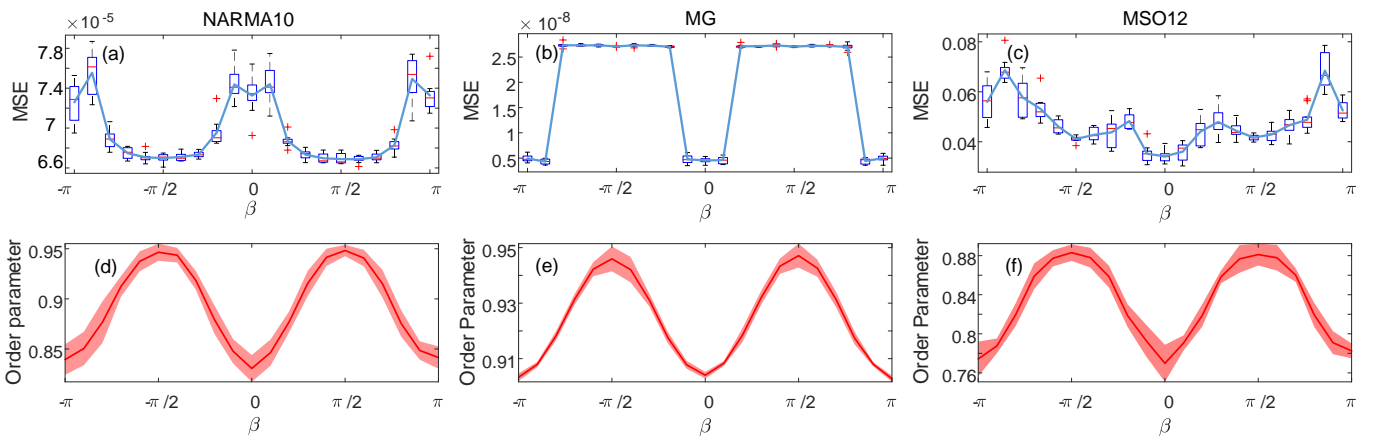

Fig. 6. The effect of character parameter $\beta$ on the performance of sad-RC. (a)–(c) Character parameter $\beta$ vs MSE on task NARMA10, MG and MSO12 respectively. Solid blue line represents the mean value of errors. (d)–(e) Character parameter $\beta$ vs order parameter.

### 3.6 Compatibility with Biological Neural Network

The effect of character parameter $\beta$ is sophisticatedly studied, and we compare the evolution of connection weights distribution in the reservoir with the development of biological neural networks in vivo.

**Character Parameter $\beta$**

In the adaptive network of phase oscillators, character parameter $\beta$ determines the feature of asymptotic states, such as symmetric, asymmetric, and chaotic. In this paper, we make a comprehensive study on the effect of character parameter $\beta$ on the performance of sad-RC.

Keeping hyperparameters intact, we vary character parameter $\beta$ from $-\pi$ to $\pi$ with a step size $\pi/12$. In each step, 50 independent simulations are conducted for a one-step ahead prediction task in NARMA10, MG, and MSO12, respectively. The boxplot of errors is displayed in Fig. 6. It is observed that the errors reach their local minimum in the vicinity of $\beta = -\pi/2, \beta = 0$ and $\beta = \pi/2$, but where the sad-RC shows its best performance may depend on a specific input. The symmetry of errors in the interval $[-\pi, \pi]$ is induced by the property of the development function in equation (2). Order parameters in Fig. 6(d)-(f) reach their maximum in $\beta = -\pi/2$ and $\beta = \pi/2$, minimum in $\beta = 0$. The results may support the assertion "the reservoir networks show their great performance in first phase transition or second phase transition".[37]

**Development of Reservoir Network**

Figure 7 shows the structures of developed reservoir networks governed by character parameter $\beta$ with different values. Even though $\beta$ can make a symmetric effect on the performance of the reservoir, the structure of reservoirs is far from symmetric about $\beta$. In Fig. 7(a), the connections of the reservoir are randomly initialized, drawn from a uniform distribution in the interval $[-1,1]$. With the constant stimulations of external inputs, the initial networks develop continually governed by specific autonomous development mechanisms. The developed networks with $\beta = -\pi/2, \beta = 0$ and $\beta = \pi/2$ are displayed in Fig. 7(b)–(d), and according to videos of the development process in Supplementary. Videos show that sharp adjustment of synaptic weights only happened in the first 3–5 steps. Although 100-step development is conducted, the distribution of the synaptic weights is brought into a steady state after the first 10 steps. The adjustment of according frequency distribution histogram of synaptic weights is invisible. This discovery confirms that our structural autonomous development mechanism enjoys a great convergence.

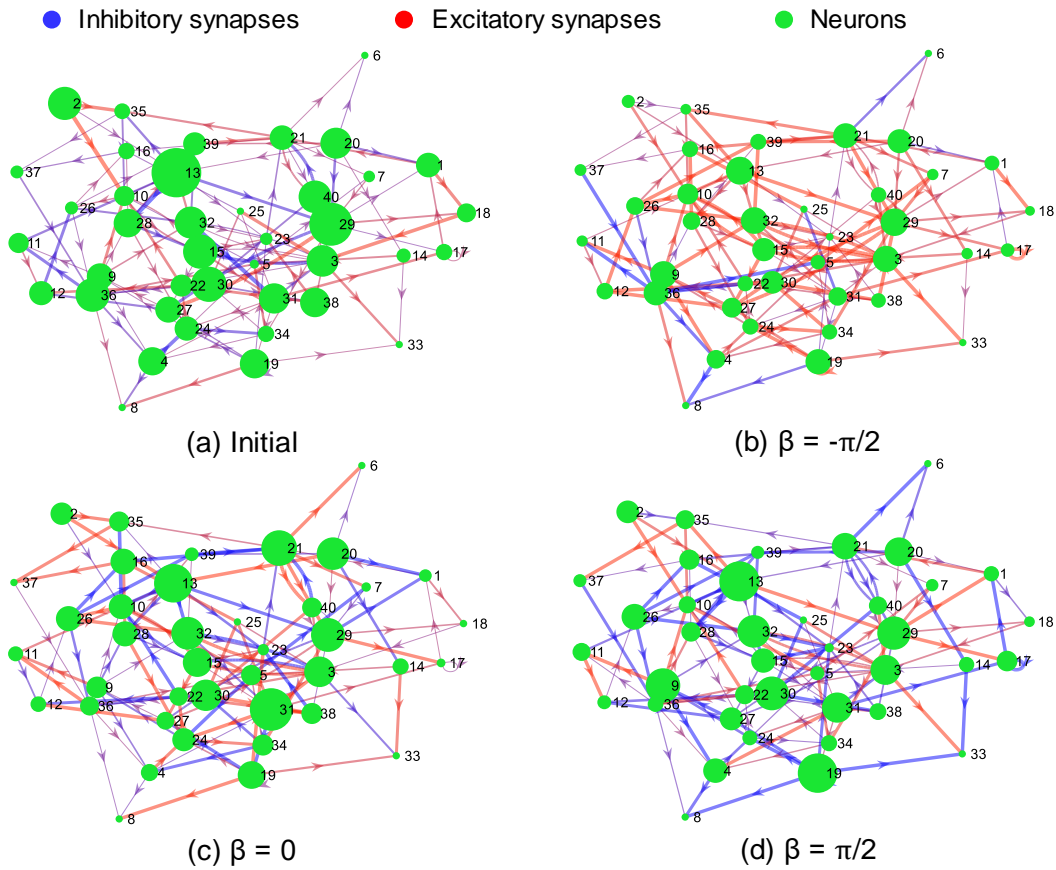

Fig. 7. Structure of reservoir network. (a) Initial network; Developed network with (b) $\beta = -\pi/2$, (c) $\beta = 0$ and (d) $\beta = \pi/2$. Red and blue color represent excitatory (positive) and inhibitory (negative) synapses respectively. Width of synapses is proportional to the absolute value of connection weights. Size of neurons is proportional to degree of them.

TABLE II. Synaptic weights distribution in reservoir. The initial distribution in the first column comes from beta distribution. Synaptic weights in developed reservoir networks are fitted by beta distribution, marked by red curve, and combinations of $a$ and $b$ are marked below each histogram.

| Initial synapses | Developed synapses with character parameter $\beta$ | | | Synapses in vivo |
|---|---|---|---|---|
| $k_{ij} \sim B(a,b)$ | $\beta = -\pi/2$ | $\beta = 0$ | $\beta = \pi/2$ | |
| 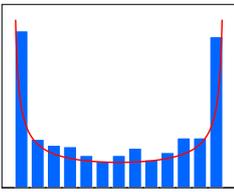 | 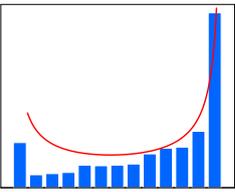 | 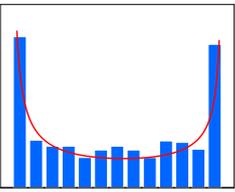 | 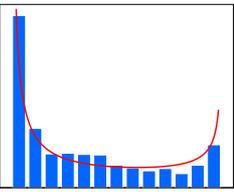 | 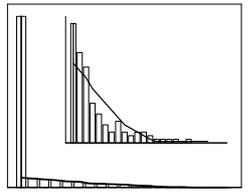 |
| (0.4, 0.4) | (0.3678, 0.2813) | (0.3054, 0.3780) | (0.1629, 0.4384) | Hebbian rule[48] |
| 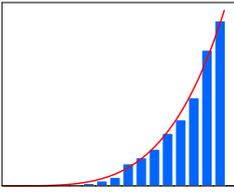 | 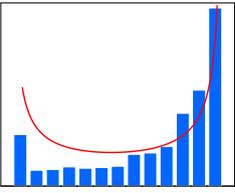 | 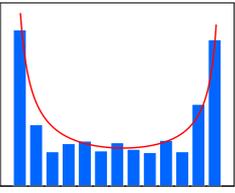 | 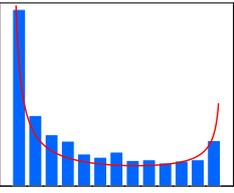 | 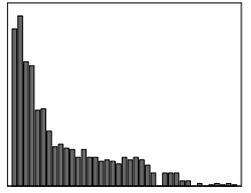 |
| (5,1) | (0.3671, 0.2844) | (0.3091, 0.3788) | (0.1646, 0.4276) | Hebbian rule[49] |
| 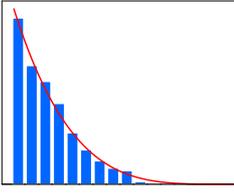 | 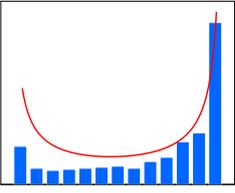 | 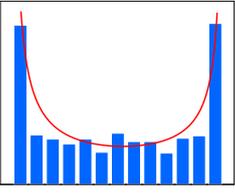 | 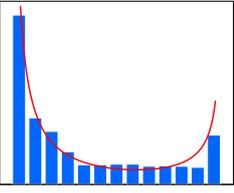 | 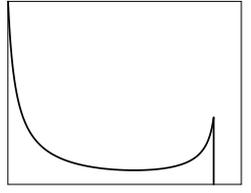 |
| (1,5) | (0.3742, 0.2812) | (0.3080, 0.3722) | (0.1614, 0.4223) | STDP[47] |
| 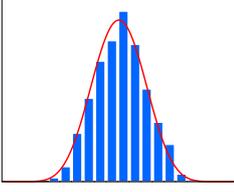 | 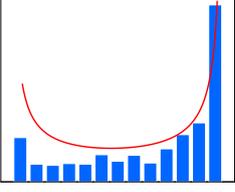 | 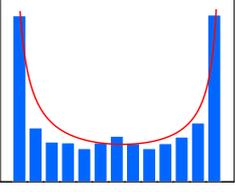 | 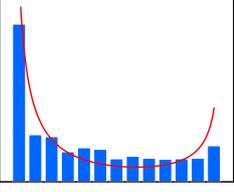 | 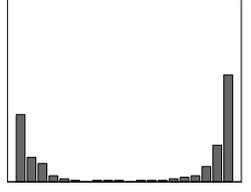 |
| (10,10) | (0.3693, 0.2845) | (0.3067, 0.3750) | (0.1608, 0.4299) | STDP[46] |
| 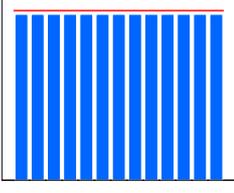 | 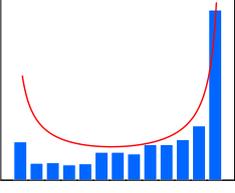 | 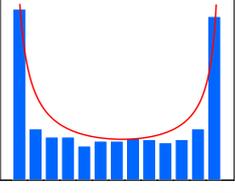 | 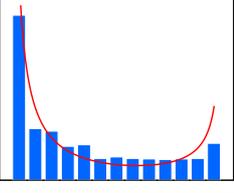 | 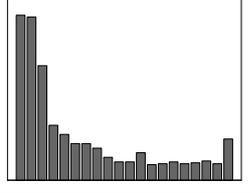 |
| (1,1) | (0.3660, 0.2822) | (0.3063, 0.3734) | (0.1605, 0.4299) | Hebbian rule[46] |
| 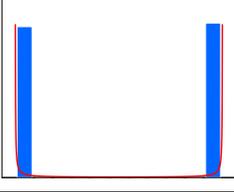 | 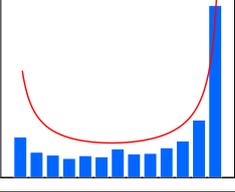 | 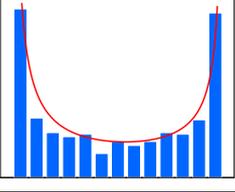 | 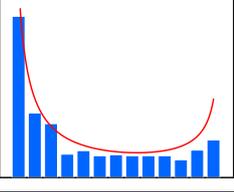 | 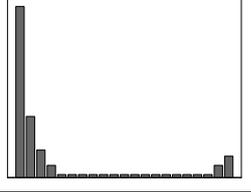 |
| (0.0001,0.0001) | (0.2430, 0.2295) | (0.2389, 0.3361) | (0.2256, 0.4801) | Hebbian rule[46] |

In the vicinity of $\beta = -\pi/2$, we have equation (2) $\sim cos(\Delta\theta)$, here $\Delta\theta = \theta_j - \theta_i$, thus, the coupling weight is increased (decreased) when the phase difference between the two oscillators is small (large). In other words, the development of synapses in the reservoir networks obeys a like-and-like (different-and-different) rule, qualitatively similar to Hebbian learning.[46] Consequently, synaptic weights converge to a stable two-cluster state, $i.e.$, positive and negative, and tend to lay more in the positive cluster with positive global coupling weights. On the contrary, synaptic weights lay more in the negative cluster with $\beta = \pi/2$. In the vicinity of $\beta = \pi/2$, the function in equation (2) is qualitatively similar to $-cos(\Delta\theta)$. The development rule of synaptic weights has the opposite effect of the like-and-like rule (Hebbian-like) in the case of a two-cluster state. We thus refer to this rule as an anti-Hebbian-like rule. Around $\beta = 0$, we have equation (2) $\sim -sin(\Delta\theta)$, and thus here, the sign of equation (2) is opposite for $\Delta\theta = \pm\Delta\theta^*$, which causes changes in the two coupling weights in opposite directions. This situation is essentially the same as that in the case of the rule for spike-timing-dependent plasticity (STDP) reported in Ref. 47. Since the property of sinusoid function in equation (2), the synaptic weights have opposite signs with the same strength. This phenomenon results in synaptic weights converging to a stable symmetric two-cluster state.

**Compare with Synaptic Weights Distribution in Vivo**

Videos in the Supplementary confirm that our model has the advantage of quick convergence. We further explore whether different initial adjacency matrices can converge to substantially identical structures. Table II shows the histogram of synaptic weights with different character parameter $\beta$. The synaptic weights come from a beta distribution $B(a,b)$ with different combinations of $a$ and $b$. Firstly, to demonstrate stable convergence of the model, reservoir networks are initialized with some representative extreme distribution as shown in the first column. It is observed that, no matter how reservoir networks are initialized, in a given character parameter, the networks will converge on a specific structure eventually. Besides, the distribution of synaptic weights of developed reservoir networks exhibits great similarity with synapses in vivo, which reassert that our model can simulate the development process of neural networks in vivo very well.

## IV. Conclusion

Based on the adaptive network of phase oscillators, we propose a structural autonomous development RC. In the washout phase of the traditional training method, the first 100 to 200 steps of neuronal states are discarded to wash out the initial neuronal state. Replaced washout phase by the autonomous development stage, the neural network can autonomously develop and adjust synaptic weights distribution with external stimulation. The dynamics of nodes and synapses in the reservoir constantly evolve together. The autonomous development mechanism makes the reservoir adapt its internal dynamics to the given task. Consequently, a stable task-tailored neural network structure is reshaped.

Simulations showed that, without elaborate design, our model maximizes its information processing capacity in a vast domain of the parameter field instead of in a narrow critical line. The benchmark memory capacity tests indicate that better prediction performance is tied to higher memory capacity.

Besides, the Kuramoto order parameter is introduced to evaluate the neuronal synchronization state in the reservoir. Results clearly indicate that the quasi-synchronous domain matched the great performance domain very well. Combined with the finding that the reservoir performs optimally when the dynamic regime is at the edge of chaos, we can infer that the sad-RC tends to evolve "the edge of chaos" state in the reservoir. Explosive synchronization is far more general in adaptive networks.[36] We presume that after autonomous development, the synchronous state in adaptive networks is disturbed by external inputs and degraded to a critical synchronous state in which the reservoir can exhibit optimal information processing capacity.

Compared with most existing RC, the sad-RC displays dominant robustness. No matter how to initialize adjacency matrices, the networks will develop into an almost identical structure in several steps. Since the autonomous development mechanism is only applied in the washout period of RC, the additional computational cost is extremely low and can even be ignored. Besides, compared with washout transient state in the reservoir, the input data is fully used to develop a specific reservoir network.

Last but not least, our model can provide a general framework for neuromorphic computing in that the development of neural networks can be easily controlled by the character parameter. The sad-RC can perform well in benchmark tests. Besides, the most important property is that the model enjoys excellent compatibility with biological neural networks, which provides an effective tool for understanding and modeling neural network development in vivo. Experiments in both representative artificial and real-life dynamic systems showed that, by adjusting the character parameter, the model could adequately simulate some development mechanisms of neural networks in vivo, such as the Hebbian rule and STDP. The autonomous development mechanism has high generality, which can be applied in RC and other artificial neural network models. The critical state of the network can be achieved easily in our model, where the networks maximize their information processing capacity. Simulations reveal that sparse reservoir networks generally work well. Sparse connections contribute to lower computational costs and more economical implementation of RC, which is conducive to developing neuromorphic hardware.

Furthermore, our model provides an efficient paradigm for understanding and modeling the development of neural networks. A unified developmental mechanism may not exist in biological neural networks. Further extending our autonomous development mechanism to multiple character parameters in multilayer networks will follow as future work.

## Acknowledgments

This study was supported in part by the China Scholarship Council under Grant 202206100025, in part by Shanghai Municipal Science and Technology Major Project No.2021SHZDZX0103, in part by Shanghai Engineering

# Appendix

**Part I** The experiment results in real-life dynamics of Sea Clutter.

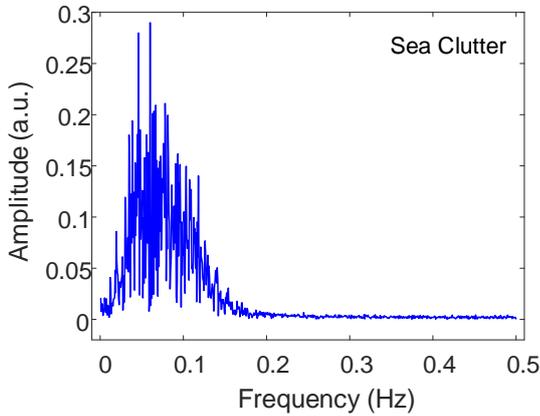

S. 1. Fast Fourier Transform results of the input signal of Sea Clutter.

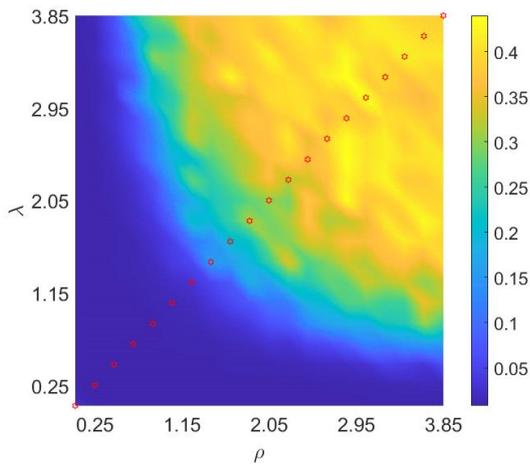

S. 2. Test error according to the global coupling strength $\lambda$ and spectral radius $\rho$ in Sea Clutter.

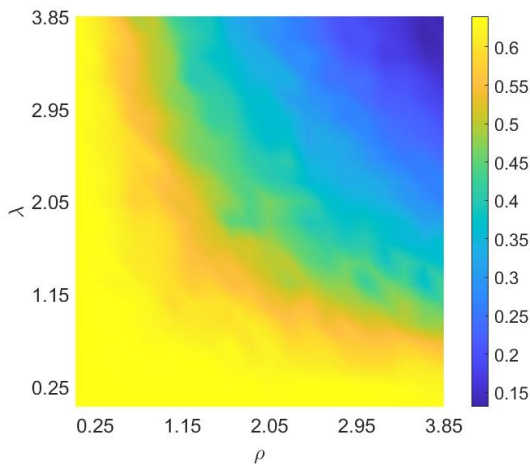

S. 3. The measure of synchrony in reservoir nodes.

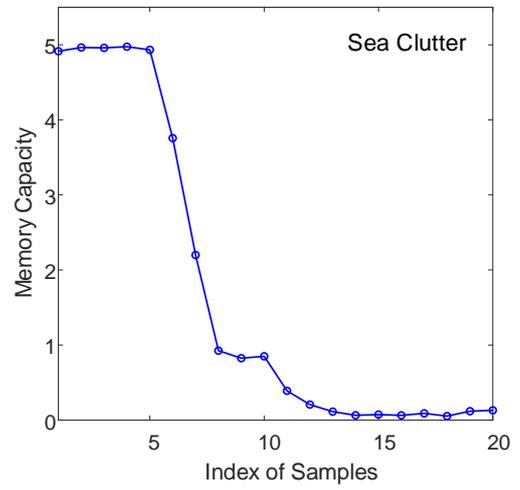

S. 4. The memory capacity of the sad-RC in Sea Clutter.

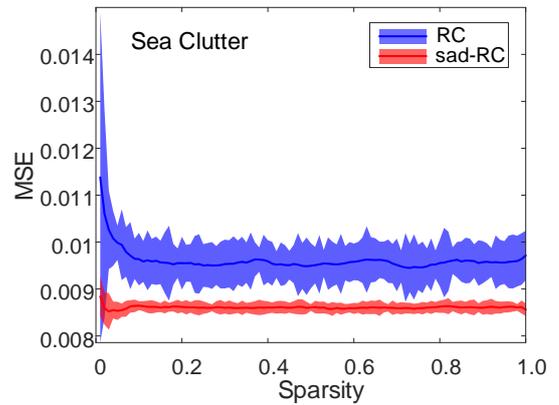

S. 5. MSE of test errors versus sparsity in the reservoir.

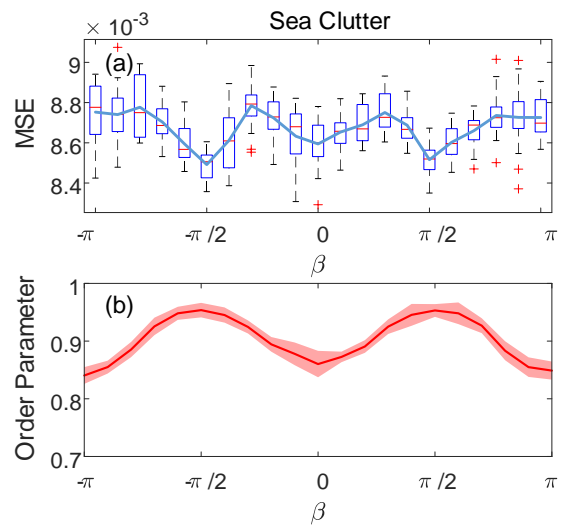

S. 6. The effect of character parameter $\beta$ on the performance of structure self-adaptive RC. (a) Character parameter $\beta$ versus MSE on task sea clutter. The solid blue line represents the mean value of errors. (b) Character parameter $\beta$ versus order parameter.

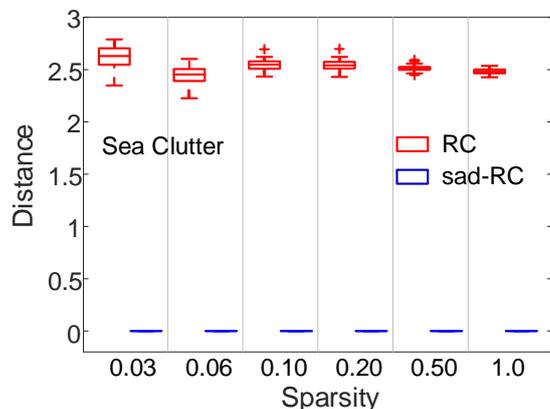

S. 7. The distance of adjacency matrices from 50 independent simulations with and without autonomous development mechanism in each sparsity, respectively.

**Part II**

The histogram of synaptic weights and neural network structure of RC with different development character parameter $\beta$. The initial synaptic weights come from a beta distribution $B(a, b)$ with different combinations of $a$ and $b$.

Videos:

(1) A_0_a10b10.mp4 and N_0_a10b10.mp4
  $\beta = 0$, $a = 10$ and $b = 10$.

(2) A_-pi2_a1b5.mp4 and N_-pi2_a1b5.mp4
  $\beta = -\pi/2$, $a = 1$ and $b = 5$.

(3) A_pi2_a10b2.mp4 and N_pi2_a10b2.mp4
  $\beta = \pi/2$, $a = 10$ and $b = 2$.